\pdfoutput=1

\documentclass[11pt]{article}

\usepackage[]{emnlp2021}

\usepackage{times}
\usepackage{latexsym}
\usepackage{booktabs}
\usepackage{url}
\usepackage{hyperref}

\usepackage[algoruled,ruled,vlined,noend]{algorithm2e}
\usepackage{enumitem}
\usepackage{array}

\newcolumntype{?}{!{\vrule width 1pt}}

\newcommand{\myparagraph}[1]{\subparagraph{\emph{#1}}\mbox{}\\}
\usepackage{array}
\usepackage{caption}
\usepackage[T1]{fontenc}

\usepackage[utf8]{inputenc}

\usepackage{microtype}

\title{Reducing Gender Bias in Machine Translation through Counterfactual Data Generation}

\author{Ranjita Naik\thanks{All authors are affiliated with Microsoft.}
\thanks{Contact author at \texttt{\scriptsize ranjitan@microsoft.com}.} \And Spencer Rarrick \And      Vishal Chowdhary
}

\begin{document}
\maketitle
\begin{abstract}

Recent advances in neural methods have led to substantial improvement in the quality of Neural Machine Translation (NMT) systems. However, these systems frequently produce translations with inaccurate gender \cite{stanovsky-etal-2019-evaluating}, which can be traced to bias in training data. 

\citet{saunders2020genderbias} tackle this problem with a handcrafted dataset containing balanced gendered profession words. By using this data to fine-tune an existing NMT model, they show that gender bias can be significantly mitigated, albeit at the expense of translation quality due to catastrophic forgetting. They recover some of the lost quality with modified training objectives or additional models at inference. We find, however, that simply supplementing the handcrafted dataset with a random sample from the base model training corpus is enough to significantly reduce the catastrophic forgetting. 

We also propose a novel domain-adaptation technique that leverages in-domain data created with the counterfactual data generation techniques proposed by \citet{zmigrod-etal-2019-counterfactual} to further improve accuracy on the WinoMT challenge test set \cite{stanovsky-etal-2019-evaluating} without significant loss in translation quality. We show its effectiveness in NMT systems from English into three morphologically rich languages – French, Spanish, and Italian.  The relevant dataset and code will be available at <Github>

\end{abstract}

\section{Introduction}

Neural Machine Translation (NMT) is now mainstream, both academically and commercially, and has become one of the biggest success stories of Deep Learning in Natural Language Processing (NLP). However, there has been a growing concern about bias in NMT systems, where models optimized to capture the statistical properties of data collected from the real world inadvertently learn or even amplify social biases found in training data. Specifically, gender bias is prevalent in widely used industrial NMT systems \cite{stanovsky-etal-2019-evaluating}.

Gender bias in NMT can be divided into two main categories of problems: (1) Translating sentences where gender is ambiguous. In this scenario, NMT systems tend to produce stereotypical gender roles in the target. For instance, while translating sentences from English to Spanish, \emph{engineer} is usually translated as being male while \emph{nurse} is usually translated as being female. (2) Translating sentences where a human can reasonably infer gender in the source from the available context. In this case, NMT output disagrees with the source gender, reflecting potential bias.

In this work, we propose solutions to address the second class of problems described above – where gender can be inferred from source context. The most prevalent approaches to mitigating gender bias in NMT models are retraining and fine-tuning. While \citet{vanmassenhove-etal-2018-getting}; \citet{DBLP:journals/corr/abs-1901-03116}; \citet{DBLP:journals/corr/abs-2010-06203} explore approaches involving retraining from scratch, \citet{costa-jussa-de-jorge-2020-fine} and \citet{saunders2020genderbias} treat gender bias as a domain-adaptation problem. Since retraining is uneconomical, we explore fine-tuning approaches to mitigate the bias. 

\citet{costa-jussa-de-jorge-2020-fine} fine-tune base NMT models on a gender balanced dataset extracted from Wikipedia. While they show an accuracy improvement of 10\% on the WinoMT pro-stereotypical subset, improvement on the anti-stereotypical subset is limited. On the other hand, \citet{saunders2020genderbias} demonstrate that by fine-tuning an existing NMT model on a handcrafted dataset containing gendered profession words, gender bias can be reduced significantly, though at the expense of translation quality due to catastrophic forgetting. They overcome that catastrophic forgetting through a regularized training technique or through inference with a lattice rescoring procedure. 

Conversely, our approach uses a subset of the training corpus itself to generate finetuning data that is in-domain. As a result, we avoid the catastrophic forgetting otherwise seen during domain adaptation. We construct our finetuning corpus by using the counterfactual data generation technique proposed by \citet{zmigrod-etal-2019-counterfactual} to convert between masculine-inflected and feminine-inflected sentences in morphologically rich target languages. We achieve 19\%, 23\%, and 21.6\% WinoMT accuracy improvements over the baseline for Italian, Spanish, and French respectively, without significant loss in general translation quality.

The advantages of our approach are three-fold
\setlist{nolistsep}
\begin{itemize}[noitemsep]
    \item Since our approach uses a subset of the in-domain training corpus to generate finetuning data, we avoid the catastrophic forgetting otherwise seen during domain adaptation.
    \item Our approach is purely data centric and therefore requires no modification to training objective or additional models at decoding. It therefore incurs no additional cost during training or inference.
    \item Using counterfactual data generation techniques, one can leverage a much more dynamic and diverse data set in model training. 
\end{itemize}

The rest of the paper is organized as follows: Section (2) covers related work. Section (3) has ethics consideration of our work. Section (4) details our data generation and fine-tuning techniques. In Section (5) we explain our experimental setup. Result and error analysis are covered in section (6)  Finally, in Section (7) we discuss future work and other concluding thoughts. 

\section{Related Work}

\citet{DBLP:journals/corr/abs-1809-02208} investigate the problem of gender bias in machine translation. They construct a templatized test set, using a list of job positions from the U.S. Bureau of Labor Statistics (BLS), in 12 gender-neutral languages and translate these sentences into English using Google Translate. They observe that Google shows a strong inclination towards male defaults, especially for fields related to science, engineering, and mathematics.

 \citet{saunders2020genderbias} approach gender bias as a domain-adaptation problem. They fine-tune a base model on a handcrafted, gender-balanced profession dataset. They show significant improvements in gender bias on WinoMT challenge set, but with a loss in general translation quality due to catastrophic forgetting. They propose two strategies for overcoming this catastrophic forgetting. Elastic Weight Consolidation (EWC), a regularized adaptation technique, involves adding a regularization term to the original loss function to maintain the general translation quality. The alternative solution, lattice rescoring, avoids forgetting by using constrained decoding to keep the translations close to the previously generated translation. These solutions require modification to the objective function or additional cost during inference. \citet{costa-jussa-de-jorge-2020-fine} fine-tune base NMT models on a gender balanced dataset extracted from Wikipedia. While they show an accuracy improvement of 10\% on the WinoMT pro-stereotypical subset, improvement on the anti-stereotypical subset is limited.

On the other hand, \citet{vanmassenhove-etal-2018-getting}; \citet{DBLP:journals/corr/abs-1901-03116}; \citet{DBLP:journals/corr/abs-2010-06203} explore approaches involving retraining from scratch.
\citet{vanmassenhove-etal-2018-getting} prepend source sentences with a tag indicating the speaker’s gender, both during training and inference. Though this doesn’t remove the gender bias from the model, it does provide some control of the gender of entities in the produced translations. \citet{DBLP:journals/corr/abs-2010-06203} train their NMT models from scratch by annotating source language words with the grammatical gender of the corresponding target words. During inference, since they do not have access to gender information of the target sentence, they use co-reference resolution tools to infer gender information from the source sentence instead. They show accuracy improvements of up to 25\% on the WinoMT challenges set. While this improvement is substantial, their approach requires annotation of the training corpus, as well as full retraining of NMT models from scratch, each of which may be prohibitively expensive for some purposes. 

\citet{saunders2021worst} approach gender errors in NMT output as a correction problem and explore methods of guiding NMT inference towards finding better gender translations. They experiment with a combination of two techniques - (1) applying gender constraints while decoding to improve nbest list gender diversity and (2) reranking nbest lists using gender features obtained automatically from the source sentence. They show WinoMT accuracy improvements similar to \citet{saunders2020genderbias}.

\section{Bias Statement}
In this work, we measure and attempt to mitigate gender bias in NMT systems. Our work only deals with scenarios where the gender can be inferred in the source. Our proposed solution involves fine-tuning a base model on a gender-balanced in-domain dataset built from the training corpus. We show substantial accuracy improvements as measured by the WinoMT test set. Both our work and the WinoMT test set are geared toward profession words, thus we may fall short in other areas. We haven't analyzed the skew in gender representation in the training data. Finally, we have only looked at bias with respect to male and female genders and our work does not address non-binary  gender identities.

\section{Gender-Balanced Data Generation }
In this section we detail the core algorithm for generating gender-balanced fine-tuning data with counterfactuals.  We generate counterfactual sentence pairs from the training data used to train our base model. The goal of this process is to identify sentences pairs that contain a masculine or feminine form of a profession animate noun and produce a modified version with the opposite gender. The modified sentence pair should be an adequate translation and the source and target should each be fluent sentences. The methodology is summarized in Algorithm 1.

\begin{algorithm}[ht]
 \SetAlgoNoLine
 \SetNoFillComment
   \textbf{Input} : {Parallel Corpus $D_{parallel}$, List of profession nouns in English $PN_{en}$} \\
    \textbf{Output} :  {$\theta_{fine-tuned}$ } \\
  \small{
  \begin{enumerate}
    \itemsep0em 
    \item $\theta_{base} \Leftarrow$ Train($D_{parallel}$),
    \item $D_{gd} \Leftarrow$ GetGenderedData($D_{parallel}$, $PN_{en}$)
    \item $D_{cf}^{src} \Leftarrow$ GenderSwapEnglish($D_{gd}^{src}$)
    \item $D_{cf}^{tgt} \Leftarrow$ GenerateTargetCounterfactuals($D_{gd}^{tgt}$)
    \item $D_{rs} \Leftarrow$ RandomSample($D_{parallel}$)
    \item $\theta_{fine-tuned} \Leftarrow$ Finetune($\theta_{base}$, ($D_{gd}$, $D_{cf}$, $D_{rs}$))
    \item Return $\theta_{fine-tuned}$
 \end{enumerate}
 }
\caption{Gender-balanced data generation and fine-tuning process  \label{algo:generate-dataset}}
\end{algorithm}

\begin{table*}[h]
\centering
\begin{tabular}{l|c|c|ccc}
\hline
& & & \multicolumn{3}{c}{\textbf{Finetuning Data}} \\
\textbf{Language Pair} & \textbf{Base Training} & \textbf{Dev} & \textbf{Gender-Balanced}(GB) & \textbf{Random} & \textbf{Handcrafted}(S\&B)\\
\hline
en $\rightarrow$  es  & 76.9M  & 869   & 413   & 200 & 388 \\
en $\rightarrow$  fr  & 134.7M & 1,160 & 1,019 & 500 & 388 \\
en $\rightarrow$  it  & 35.1M  & 938   & 1,041 & 500 & 388 \\ 
\hline
\end{tabular}
\caption{Corpus sizes (sentence pairs)}
\label{table:Corpus_sizes}
\end{table*}

\subsection{In-Domain Data Generation}
\myparagraph{Selecting Gendered Sentences}
\label{sec:Selecting_Gendered_Sentences}
We begin the finetuning data generation process by selecting a subset of the base model training corpus to use for counterfactual data generation. To generate high quality data which also works with the counterfactual generation tools, we reject sentences that do not meet the following criteria:
\begin{itemize}
  \item \emph{Length}: Maximum of 20 whitespace-delimlited tokens.
  \item \emph{Length Ratio}: Ratio between tokens on each side does not exceed 3.
  \item \emph{Animacy}: English sentence must contain exactly one gendered pronoun and one profession noun from the list extracted from the handcrafted dataset. Additionally, we use Stanza (\citet{qi2020stanza}) to check that the Part-of-speech (POS) of the matched profession word is noun, as some have adjective senses as well.
  \item \emph{Wellformedness}: English side must begin with a capital letter and end with a punctuation mark.
  \item \emph{Proper Nouns}: English side contains no tokens tagged as proper nouns by Stanza.
\end{itemize}
\subsection{Counterfactual Data Generation}
\paragraph{\emph{Generating Counterfactuals for Source Data:}}

We produce a gender-swapped version of each English sentence in the set extracted in  \ref{sec:Selecting_Gendered_Sentences} by replacing the pronoun in our replacement set with its opposite-gendered counterpart. \emph{Her} can be swapped to either \emph{him} or \emph{his}, and we use the Spacy POS tagger to disambiguate. \emph{His} is always swapped to \emph{her}, as cases where it should be swapped to \emph{hers} are sufficiently rare. 

\paragraph{\emph{Generating Counterfactuals for Target Data:}}
In morphologically rich languages, this generally requires changing the form for more words than just the identified animate noun in the target language. For example:

\hspace{1cm}\emph{{\color{red}Le soldat} {\color{teal}allemand} est très {\color{blue}content}.}

\hspace{1cm}\emph{{\color{red}La soldate} {\color{teal}allemande} est très {\color{blue}contente}.}

We leverage \citet{zmigrod-etal-2019-counterfactual} for generating the counterfactual data for Spanish, French and Italian sentences.  This begins by using Stanza \citet{qi2020stanza} to tokenize and parse the target sentences. We then use Udify \citet{kondratyuk201975} to add universal dependency features such as number and gender to tokens in the parse tree.

Each target sentence has been preselected to contain exactly one profession animate noun from our animacy list. We mark this animate noun to be gender-swapped and apply a Markov Random Field (MRF) model as proposed by \citet{zmigrod-etal-2019-counterfactual} to identify additional tokens that also need to have their gender changed so that the generated counterfactual sentence has correct gender agreement. For French and Spanish, we use the models provided by \citet{zmigrod-etal-2019-counterfactual}, and for Italian we trained a model from treebank data\footnote{\url{https://universaldependencies.org/\#download}}. We found that for French and Italian, the MRF model rarely marked determiners for reinflection, so we additionally marked any determiner token whose head is the selected animate noun.

Next we reinflect the lemmata of the tokens marked in the previous step to their new forms. During initial experimentation we used a Morphological Inflection (MI) model to reinflect all identified tokens, but we found that it had low accuracy on determiners and often left the profession animate nouns themselves unchanged from their original form. Furthermore, in Italian and French the correct form for some determiners depends on sentence context and therefore cannot be correctly predicted by an MI model without access to that context. We determined that using dictionary lookup for determiners and nouns in the profession animacy list produces better results. We fall back to the MI model for words not covered in the dictionary, which most often were adjectives modifying the animate noun. 
We apply a set of hand-crafted rules to form contractions between prepositions and articles as necessary for each language. For example, in Spanish \emph{de el} (glossed as \emph{of the} in English) is contracted to \emph{del}. We then detokenize the sentences.

\paragraph{\emph{Generating counterfactuals through forward and backtranslation:}} During early testing we explored alternative approaches for generating each side of the data in the gender-balanced dataset. We backtranslated the target counterfactual to produce a corresponding English sentence. Likewise, we forward translated the gender-swapped source sentences in English to generate counterfactuals in the target language. However, we found the results of this alternative method to be inferior. One possible explanation for this inferior quality is that both forward and back translation systems have the same gender bias problem that we are trying to solve, and so it was not able to produce translations of the needed quality.

\subsection{Selecting Random Dataset}We also sample another subset of the training corpus to use as neutral data. Our aim is to select a dataset which has distribution more similar to the original base model training corpus. This data may or may not contain any gendered pronouns or words from our profession animate noun list. Our hypothesis is that including this data along with the handcrafted or counterfactual datasets during fine tuning will help mitigate the catastrophic forgetting that is observed when training on those datasets alone.

We apply the following filters  from \ref{sec:Selecting_Gendered_Sentences} to the base model training corpus before randomly sampling: \emph{Length (maximum of 100 tokens)}, \emph{Length Ratio} and \emph{Wellformedness}.

\subsection{Handcrafted Profession Dataset}
We utilize and extend the handcrafted profession dataset developed by \citet{saunders2020genderbias} in this work. This data set consists of 388 English sentences human-translated into each target language of interest. Each English sentence consists of a profession word from a list of 194, embedded into a masculine or feminine version of the same template. For example:

 \hspace{1cm}\emph{The logistician finished his work.}
 
 \hspace{1cm}\emph{The logistician finished her work.}
 
We use the Spanish translations from \citet{saunders2020genderbias} and have additionally translated the English data into French and Italian.

\section{Experiments}

\subsection{Languages and Data}
We evaluate our approach on three language pairs in a high-resource setting – English $\rightarrow$  French (\textbf{en $\rightarrow$  fr}), English $\rightarrow$  Spanish (\textbf{en $\rightarrow$  es}) and English $\rightarrow$  Italian (\textbf{en $\rightarrow$  it}).  The WinoMT framework supports the evaluation of gender bias in translations from English into these morphologically rich target languages from romance language family.  
  
Following the procedure described in \ref{sec:Selecting_Gendered_Sentences}, we sample at most 10 training examples for each word in our profession animate noun list and generate counterfactuals for those sentences. We then remove any sentence pair for which there is no difference between the counterfactual and the original target sentence.

We then create sentence pairs from each gender-swapped English sentence and corresponding counterfactual target sentence. We add to this the original non-counterfactual sentence pairs so that for each sentence pair, we have a balanced masculine and feminine version. Combining original and counterfactual data in this way yields two interesting benefits. First, each profession word will now be balanced in this dataset between male and female versions, which should guide the model towards a gender-balanced state rather than overshooting. Second, because the sentential context is identical in the male and female versions, we hope not to teach the model to erroneously associate male and female versions with unrelated contextual features. 

Table \ref{table:Corpus_sizes} summarizes the corpus statistics for train, validation and fine-tune datasets for the three language pairs. We use IWSLT dataset for validation and WMT datasets for testing.

\subsection{Training and inference}

Our base NMT models are transformers\citet{vaswani2017attention} with RNN-based decoder with SSRU \citet{kim-etal-2019-research}, implemented in the Marian Toolkit \citet{junczysdowmunt2018marian}. We use 6 layers in the encoder and decoder, 8 attention heads, and 2048-dimension feed-forward layer with RELU activation. We apply a dropout of 0.1 between transformer layers. The embedding dimension is 512 and we tie target embeddings and output embeddings.

We learn a joint vocabulary with 32K tokens, computed with SentencePiece on the training data. We set label smoothing to 0.1. We optimize the model parameters with SGD(adam), with a learning rate schedule of (learning rate, warmup)= (0.0002, 8000). We train baselines with AfterBatches set to 1000000 and early stopping on validation set cross-entropy-mean-word set to 10. We decode with beam size=4.

During fine-tuning we use the vocabulary of the base model. We optimize the model parameters with SGD(adam), with a learning rate schedule of (learning rate, warmup)= (5e-05, 100). We set AfterBatches to 4000 with early stopping on validation BLEU set to 20. 

\begin{table*}[h]
\centering
\setlength{\tabcolsep}{2pt}
\begin{tabular}{l?c|c|c|c|c?c|c|c|c|c?c|c|c|c|c}
\toprule
\multicolumn{1}{l}{} & \multicolumn{5}{c}{\textbf{Spanish}} & \multicolumn{5}{c}{\textbf{French}} & \multicolumn{5}{c}{\textbf{Italian}}  \\
\midrule

\textbf{System} & Acc & Pro & Anti & $\Delta$S & $\Delta$G & Acc & Pro & Anti & $\Delta$S & $\Delta$G & Acc & Pro & Anti & $\Delta$S & $\Delta$G \\ \hline
Baseline        & 55.8 & 69.9 & 52.0 & 17.9 & 18.3 & 53.5 & 71.9 & 41.4 & 30.8 & 10.3 & 41.2 & 51.8 & 34.9 & 16.9 & 32.2 \\ \hline
Handcrafted(S\&B)         & 76.8 & 89.6 & 77.6 & \textbf{12.0} & -2.8 & \textbf{78.2} & \textbf{92.2} & \textbf{76.1} & \textbf{16.1} & -6.6 & 56.8 & 67.2 & 53.2 & 14.0 & 3.8  \\
S\&B+Random     & 75.9 & 88.8 & 76.3 & 12.5 & \textbf{-2.1} & 72.2 & 88.5 & 66.5 & 22.0 & \textbf{-6.2} & 55.8 & 66.6 & 51.3 & 15.3 & 4.2  \\ \hline
Gender-Balanced(GB)             & 77.2 & 92.3 & 75.4 & 16.9 & -4.4 & 70.6 & 86.0 & 64.0 & 22.0 & -7.6 & 57.3 & 64.8 & 56.6 & \textbf{8.2}  & 4.4  \\
GB+Random       & 77.2 & 92.3 & 75.3 & 17.0 & -4.4 & 73.4 & 86.4 & 69.9 & 16.5 & -7.3 & 58.0 & 65.9 & 56.6 & 9.3  & 3.3  \\
GB+Random+S\&B & \textbf{78.8} & \textbf{93.0} & \textbf{78.7} & 14.3 & -4.5 & 75.1 & 90.0 & 71.1 & 18.9 & -7.1 & \textbf{60.2} & \textbf{68.8} & \textbf{59.2} & 9.6  & \textbf{-0.2} \\ \hline

\end{tabular}
\caption{WinoMT Accuracy - Best scores in each column are in bold: highest numbers for accuracy measures and lowest absolute value for $\Delta$S and $\Delta$G.Though the S\&B system in French has the best accuracy, Pro, Anti, and $\Delta$S, it suffers from a drop in BLEU of up to 1 point. }
\label{table:WinoMT_Accuracy}
\end{table*}

\begin{table*}[ht]
\vspace*{1 cm}
\centering
\setlength{\tabcolsep}{2pt}
\begin{tabular}{l?cccc?cccc?cc}
\toprule
\multicolumn{1}{l}{} & \multicolumn{4}{c}{\textbf{Spanish}} & \multicolumn{4}{c}{\textbf{French}} & \multicolumn{2}{c}{\textbf{Italian}}  \\
\midrule
\textbf{System}      & wmt09 & wmt10 & wmt13 & wmt19 & wmt09 & wmt10 & wmt11 & wmt15 & wmt09 & wmt19 \\ \hline
Baseline        & 30.6     & 38.2     & 35.5     & 41.3     & 30.4     & 34.5     & 34.9     & 40.2     & 32.3     & 37.0     \\ \hline
Handcrafted(S\&B)            & 30.4     & 37.8     & 34.9     & 40.3     & 29.3     & 33.3     & 34.2     & 39.0     & 31.8     & 36.2     \\
S\&B+Random     & 30.5     & 38.2     & 35.5     & 41.0     & 30.0     & 34.2     & 34.7     & 39.6     & \textbf{32.3}     & \textbf{37.5}     \\ \hline
Gender-Balanced(GB)              & 30.4     & 37.9     & 35.5     & 41.3     & \textbf{30.3}     & \textbf{34.2}     & \textbf{34.9}     & \textbf{40.1}     & 32.3     & 36.8     \\
GB+Random       & 30.4     & 37.9     & 35.5     & 41.3     & 30.1     & 34.1     & 34.6     & 39.7     & 32.4     & 37.1     \\
GB+Random+S\&B  & \textbf{30.5}     & \textbf{38.2}     & \textbf{35.6}     & \textbf{41.3}     & 30.1     & 34.1     & 34.6     & 39.8     & 32.0     & 36.9     \\ \hline
\end{tabular}
\caption{Translation Quality (BLEU) - The fine-tuned system with best average BLEU across test sets is in bold.}
\label{table:Translation_Quality}
\end{table*}

\section{Results}

\subsection{WinoMT Challenge Set and Evaluation Metrics }
Following other recent work, we evaluate our NMT systems on WinoMT \cite{stanovsky-etal-2019-evaluating} challenge set to quantify gender bias. This dataset consists of English sentences, each containing two animate nouns, one of which is coreferent with a gendered pronoun. From context within the sentence, a human can clearly determine which animate noun is coreferent, and thus the gender of the person described by that noun. By checking how often an MT system produces a translation of the correct gender for that animate noun, we can determine to what extent the system relies on gender stereotypes as opposed to relevant context. Following are key metrics:

\textbf{Accuracy (Acc)} – Percentage of translations with correct gender for the primary entity.

\textbf{Pro} – Accuracy on pro-stereotypical subset (e.g. female nurse).

\textbf{Anti} – Accuracy on anti-stereotypical subset (e.g. male receptionist).

\textbf{$\Delta$S} – Difference in accuracy between pro-stereotypical and anti-stereotypical gender role assignments. Higher value indicates that models are better at translating gender when it aligns with stereotypical gender roles.

\textbf{$\Delta$G} – Difference between male and female F1 scores. Higher positive value indicates that the system does a better job in translating male entities. 

\begin{table*}[h]
\centering
\begin{tabular}{|ll|}

\hline
Source:  & {\color{red}The hairdresser} shouted at {\color{teal}the driver} because {\color{brown}she} crashed the car.  \\
Baseline MT: & {\color{red}Le coiffeur} a crié {\color{teal}au conducteur} parce qu{\color{brown}’elle} a écrasé la voiture. \\
GB+Random+S\&B:  & {\color{red}La coiffeuse} a crié à {\color{teal}la conductrice} parce qu{\color{brown}’il} a percuté la voiture. \\

S\&B+Random: & {\color{red}La coiffeuse} a crié à {\color{teal}la conductrice} parce qu{\color{brown}’elle} a écrasé la voiture. \\
\hline

\end{tabular}
\caption{Translation of English source into French by various systems}
\label{table:Translation_English_to_French}
\vspace*{4mm}
\end{table*}

\begin{table*}[h]
\centering
\begin{tabular}{|ll|}
\hline
Original English: & {\color{red}He} didn't wait around for {\color{teal} the reporter } to come calling. \\
Genderswapped English: &  {\color{red}She} didn't wait around for {\color{teal} the reporter } to come calling. \\
& \\
Original French: & {\color{red} Il} n'a pas attendu que {\color{teal} le journaliste} l'appelle. \\
Counterfactual French: &  {\color{red} Il} n'a pas attendu que {\color{teal} la journaliste} l'appelle. \\
\hline
\end{tabular}
\caption{Example source and target counterfactuals with harmful gender swapping on English side}
\label{table:Example_source_and_target_counterfactuals}
\vspace*{4mm}
\end{table*}

\begin{table*}[!h]
\centering
\begin{tabular}{|ll|}

\hline
Original English: & If {\color{red} the client} does not like the photograph,  {\color{teal} he } pays nothing. \\
Genderswapped English: & If {\color{red} the client} does not like the photograph,  {\color{teal} she } pays nothing. \\
& \\
Original French: & Si {\color{red} le client } n'aime pas la photographie,  {\color{teal} il } ne paie rien. \\

Counterfactual French: & Si {\color{red} la cliente } n'aime pas la photographie,  {\color{teal} il } ne paie rien. \\
\hline

\end{tabular}
\caption{Illustration of inaccurate counterfactual generation by the MRF model}
\label{table:Inaccurate_MRF_model}
\end{table*}

\subsection{Result Analysis}
We fine-tuned our base models on various combinations of the three datasets described in Section 4: (1) the handcrafted dataset from \citet{saunders2020genderbias} (S\&B), (2) a random sample of the base model training data (Random), and the gender-balanced in-domain dataset we describe in section 4.2 (GB). We report WinoMT metrics in Table \ref{table:WinoMT_Accuracy} and BLEU scores for each system are presented in Table \ref{table:Translation_Quality}.

Our base models achieve comparable accuracy,  $\Delta$G, and $\Delta$S to the best performing commercial translation systems reported in \cite{stanovsky-etal-2019-evaluating}. The WinoMT scores for the base models show that they are heavily gender-biased. High positive values of  $\Delta$S indicate much higher gender accuracy on sentences with pro-stereotypical assignments than anti-stereotypical ones. 

Fine-tuning on S\&B alone yields substantial improvements on WinoMT accuracy relative to the baseline, up to 25\%  for French, as well as reducing  $\Delta$G and  $\Delta$S dramatically. However, catastrophic forgetting leads to a consistent drop in BLEU of up to around 1 point, which is significant. This was also observed in original experiments in \citet{saunders2020genderbias}.

By supplementing S\&B with our random sample dataset (S\&B+Random), we recover most of the BLEU degradation from S\&B while retaining the bulk of the WinoMT accuracy gains, though with an accuracy gap of 6\% remaining on French. This demonstrates that the method is a suitable alternative for their more complicated and costly approaches of EWC and lattice rescoring.

Fine-tuning on our gender-balanced dataset (GB) achieves \emph{better} WinoMT accuracy than S\&B and S\&B+Random for Spanish and Italian with only minimal loss in BLEU relative to base models. The French system has slightly lower accuracy than S\&B , though with much less BLEU degradation and still 17\% improved accuracy over the baseline. Further adding random data (GB+random) had almost no effect on Spanish and a slight improvement in accuracy and BLEU for Italian. Because GB and Random are both in-domain, the additional benefit of including Random appears to be less than for adding it to S\&B. For French, we see a slight improvement in BLEU and slight regression in WinoMT accuracy.

Finally, our models finetuned on all three datasets combined (GB+Random+S\&B) show the strongest WinoMT accuracy among systems with acceptable levels of BLEU loss relative to base models. These systems show accuracy improvements of 19\%, 23\% and 21.6\% over the baseline for Italian, Spanish, and French respectively, as well as substantial improvement in  $\Delta$S for all the language pairs. Though accuracy for the S\&B French model is slightly higher, that system shows a drop in BLEU of around 1 point relative to the base model, which may be considered unacceptably large.

\subsection{Error Analysis}

\textbf{en $\rightarrow$  fr} models finetuned on data including our GB dataset exhibit a problematic pattern of mistranslating the English word \emph{she} into \emph{il (he)}, despite large gains in predicting correct gender of profession nouns themselves. This pattern is observed in both the WinoMT output and WMT test sets, but is not seen in the base model output or finetuning experiments that exclude the GB dataset.

For example, in Table \ref{table:Translation_English_to_French}, \emph{she} is translated to \emph{il} in the GB+Random+S\&B translation, despite both profession nouns in the sentence being changed to feminine forms (WinoMT counts this sentence correct if the translation of \emph{the driver/la conductrice} is feminine, but \emph{the hairdresser/la coiffeuse} has also been changed). In the base model and S\&B+Random translations, \emph{she} is correctly translated to \emph{elle}.

Examining the GB dataset, we notice several examples that may be contributing to this pattern. In Table \ref{table:Example_source_and_target_counterfactuals}, we see an example where the English pronoun refers to a different individual from the profession noun that we modify in the counterfactual, and our method of gender swapping English introduces an erroneous mapping between \emph{she} and \emph{il}. In Table \ref{table:Inaccurate_MRF_model} we see an example where the pronoun and profession noun refer to the same individual, but the pronoun is not identified by our MRF model. Hence, it does not get changed to feminine on the French side, but does in the English side due to the simple gender swapping rule.

We do not see the same error in our Spanish or Italian output. One likely reason for this is that subject pronouns are frequently elided in those two languages, so far fewer erroneous mappings are introduced. The model is also more able to hide uncertainty about correct pronoun gender by simply omitting subject pronouns. 

We also do not see the reverse pattern, i.e. \emph{he} translating to \emph{elle (she)}. This may be because the sampled gender dataset contains about 3 times as many instances of \emph{he} as \emph{she}. The gender swapping process produces many more \emph{she/il} mappings.

We also see another interesting pattern in Table \ref{table:Translation_English_to_French} – In the base model output both profession nouns are masculine, but in both finetuned models, the gender of both has changed to feminine, matching the pronoun, \emph{she}. This suggests that the models may not be learning to solve the coreference resolution problem so much as simply conditioning on the gender of any present pronoun. This pattern is observed frequently in all three of our tested languages.

\section{Conclusion}

In this work we demonstrate two fine-tuning-based approaches for mitigating gender bias in NMT in scenarios where gender is clear in the source. We show substantial improvements in WinoMT accuracy for three language pairs without significant degradation in BLEU.

First, we extend the approach described in \citet{saunders2020genderbias} by adding a random in-domain sample to their handcrafted profession dataset before finetuning. We show that this simple method is sufficient to minimize catastrophic forgetting and provides an attractive alternative to their potentially more complicated proposals of EWC regularization and lattice rescoring.

Next, we adapt the counterfactual-data-generation technique of \citet{zmigrod-etal-2019-counterfactual} to an NMT setting to synthesize a gender-balanced in-domain dataset for finetuning. This data-centric approach constructs the finetuning data from the original training corpus, and therefore keeps the model focused on the original domain and incurs no additional cost during training or inference

Among various interesting directions for future work, we would like to extend our techniques for addressing non-binary gender. Also, our current techniques work well in simple sentences when only a single individual is mentioned, such as \emph{“The doctor finished his work for the day”.} When multiple people are mentioned in a single sentence this method can produce sentence pairs where different entities get gender-swapped in the source and target. We plan to explore using coreference resolution systems to filter out incorrect sentence pairs in our data selection algorithm.

\bibliography{anthology,custom}

\begin{thebibliography}{14}
\expandafter\ifx\csname natexlab\endcsname\relax\def\natexlab#1{#1}\fi

\bibitem[{Costa-juss{\`a} and de~Jorge(2020)}]{costa-jussa-de-jorge-2020-fine}
Marta~R. Costa-juss{\`a} and Adri{\`a} de~Jorge. 2020.
\newblock \href {https://www.aclweb.org/anthology/2020.gebnlp-1.3} {Fine-tuning
  neural machine translation on gender-balanced datasets}.
\newblock In \emph{Proceedings of the Second Workshop on Gender Bias in Natural
  Language Processing}, pages 26--34, Barcelona, Spain (Online). Association
  for Computational Linguistics.

\bibitem[{Font and
  Costa{-}juss{\`{a}}(2019)}]{DBLP:journals/corr/abs-1901-03116}
Joel~Escud{\'{e}} Font and Marta~R. Costa{-}juss{\`{a}}. 2019.
\newblock \href {http://arxiv.org/abs/1901.03116} {Equalizing gender biases in
  neural machine translation with word embeddings techniques}.
\newblock \emph{CoRR}, abs/1901.03116.

\bibitem[{Junczys-Dowmunt et~al.(2018)Junczys-Dowmunt, Heafield, Hoang,
  Grundkiewicz, and Aue}]{junczysdowmunt2018marian}
Marcin Junczys-Dowmunt, Kenneth Heafield, Hieu Hoang, Roman Grundkiewicz, and
  Anthony Aue. 2018.
\newblock \href {http://arxiv.org/abs/1805.12096} {Marian: Cost-effective
  high-quality neural machine translation in c++}.

\bibitem[{Kim et~al.(2019)Kim, Junczys-Dowmunt, Hassan, Fikri~Aji, Heafield,
  Grundkiewicz, and Bogoychev}]{kim-etal-2019-research}
Young~Jin Kim, Marcin Junczys-Dowmunt, Hany Hassan, Alham Fikri~Aji, Kenneth
  Heafield, Roman Grundkiewicz, and Nikolay Bogoychev. 2019.
\newblock \href {https://doi.org/10.18653/v1/D19-5632} {From research to
  production and back: Ludicrously fast neural machine translation}.
\newblock In \emph{Proceedings of the 3rd Workshop on Neural Generation and
  Translation}, pages 280--288, Hong Kong. Association for Computational
  Linguistics.

\bibitem[{Kondratyuk and Straka(2019)}]{kondratyuk201975}
Dan Kondratyuk and Milan Straka. 2019.
\newblock \href {http://arxiv.org/abs/1904.02099} {75 languages, 1 model:
  Parsing universal dependencies universally}.

\bibitem[{Prates et~al.(2018)Prates, Avelar, and
  Lamb}]{DBLP:journals/corr/abs-1809-02208}
Marcelo O.~R. Prates, Pedro H.~C. Avelar, and Lu{\'{\i}}s~C. Lamb. 2018.
\newblock \href {http://arxiv.org/abs/1809.02208} {Assessing gender bias in
  machine translation - {A} case study with google translate}.
\newblock \emph{CoRR}, abs/1809.02208.

\bibitem[{Qi et~al.(2020)Qi, Zhang, Zhang, Bolton, and Manning}]{qi2020stanza}
Peng Qi, Yuhao Zhang, Yuhui Zhang, Jason Bolton, and Christopher~D. Manning.
  2020.
\newblock \href {http://arxiv.org/abs/2003.07082} {Stanza: A python natural
  language processing toolkit for many human languages}.

\bibitem[{Saunders and Byrne(2020)}]{saunders2020genderbias}
Danielle Saunders and Bill Byrne. 2020.
\newblock Reducing gender bias in neural machine translation as a domain
  adaptation problem.
\newblock In \emph{ACL}. Association for Computational Linguistics.

\bibitem[{Saunders et~al.(2021)Saunders, Sallis, and Byrne}]{saunders2021worst}
Danielle Saunders, Rosie Sallis, and Bill Byrne. 2021.
\newblock \href {http://arxiv.org/abs/2104.07429} {First the worst: Finding
  better gender translations during beam search}.

\bibitem[{Stafanovics et~al.(2020)Stafanovics, Bergmanis, and
  Pinnis}]{DBLP:journals/corr/abs-2010-06203}
Arturs Stafanovics, Toms Bergmanis, and Marcis Pinnis. 2020.
\newblock \href {http://arxiv.org/abs/2010.06203} {Mitigating gender bias in
  machine translation with target gender annotations}.
\newblock \emph{CoRR}, abs/2010.06203.

\bibitem[{Stanovsky et~al.(2019)Stanovsky, Smith, and
  Zettlemoyer}]{stanovsky-etal-2019-evaluating}
Gabriel Stanovsky, Noah~A. Smith, and Luke Zettlemoyer. 2019.
\newblock \href {https://doi.org/10.18653/v1/P19-1164} {Evaluating gender bias
  in machine translation}.
\newblock In \emph{Proceedings of the 57th Annual Meeting of the Association
  for Computational Linguistics}, pages 1679--1684, Florence, Italy.
  Association for Computational Linguistics.

\bibitem[{Vanmassenhove et~al.(2018)Vanmassenhove, Hardmeier, and
  Way}]{vanmassenhove-etal-2018-getting}
Eva Vanmassenhove, Christian Hardmeier, and Andy Way. 2018.
\newblock \href {https://doi.org/10.18653/v1/D18-1334} {Getting gender right in
  neural machine translation}.
\newblock In \emph{Proceedings of the 2018 Conference on Empirical Methods in
  Natural Language Processing}, pages 3003--3008, Brussels, Belgium.
  Association for Computational Linguistics.

\bibitem[{Vaswani et~al.(2017)Vaswani, Shazeer, Parmar, Uszkoreit, Jones,
  Gomez, Kaiser, and Polosukhin}]{vaswani2017attention}
Ashish Vaswani, Noam Shazeer, Niki Parmar, Jakob Uszkoreit, Llion Jones,
  Aidan~N. Gomez, Lukasz Kaiser, and Illia Polosukhin. 2017.
\newblock \href {http://arxiv.org/abs/1706.03762} {Attention is all you need}.

\bibitem[{Zmigrod et~al.(2019)Zmigrod, Mielke, Wallach, and
  Cotterell}]{zmigrod-etal-2019-counterfactual}
Ran Zmigrod, Sabrina~J. Mielke, Hanna Wallach, and Ryan Cotterell. 2019.
\newblock \href {https://doi.org/10.18653/v1/P19-1161} {Counterfactual data
  augmentation for mitigating gender stereotypes in languages with rich
  morphology}.
\newblock In \emph{Proceedings of the 57th Annual Meeting of the Association
  for Computational Linguistics}, pages 1651--1661, Florence, Italy.
  Association for Computational Linguistics.

\end{thebibliography}
\bibliographystyle{acl_natbib}

\end{document}